\documentclass{article}

\usepackage{times}
\usepackage{graphicx} % more modern
\usepackage{subfigure}

\usepackage{natbib}

\usepackage{algorithm}
\usepackage{algorithmic}

\usepackage{hyperref}

\usepackage[arxiv]{icml2016}

\usepackage{amsmath}
\usepackage{amssymb}
\usepackage{amsthm}
\usepackage{amsfonts}
\newtheorem{defn}{Definition}
\newtheorem{theorem}{Theorem}

\icmltitlerunning{Coupled Compound Poisson Factorization}

\begin{document}

\twocolumn[
\icmltitle{Coupled Compound Poisson Factorization}

\icmlauthor{Mehmet E. Basbug}{mehmetbasbug@yahoo.com}
\icmladdress{Princeton University,
             35 Olden St., Princeton, NJ 08540 USA}
\icmlauthor{Barbara E. Engelhardt}{bee@princeton.edu}
\icmladdress{Princeton University,
             35 Olden St., Princeton, NJ 08540 USA}
\icmlkeywords{compound poisson, factorization, variational inference}

\vskip 0.3in
]

\begin{abstract}
    We present a general framework, the coupled compound Poisson factorization (CCPF), to capture the missing-data mechanism in extremely sparse data sets by coupling a hierarchical Poisson factorization with an arbitrary data-generating model. We derive a stochastic variational inference algorithm for the resulting model and, as examples of our framework, implement three different data-generating models---a mixture model, linear regression, and factor analysis---to robustly model non-random missing data in the context of clustering, prediction, and matrix factorization. In all three cases, we test our framework against models that ignore the missing-data mechanism on large scale studies with non-random missing data,
    and we show that explicitly modeling the missing-data mechanism substantially improves the quality of the results, as measured using data log likelihood on a held-out test set.
\end{abstract}

\section{Introduction}
\label{intro}

The statistical theory of missing data developed by \citet{little2014statistical} starts with an important distinction between a missing-data pattern and a missing-data mechanism. A missing-data pattern is an indicator matrix, $M$, that describes which matrix values are missing. A missing-data mechanism, on the other hand, captures the relationship between the missing-data pattern and the data
generating model. When the distribution of $M$, the \emph{missingness-encoding model}, does not depend on the observed data, $Y_{obs}$, or the missing data, $Y_{mis}$, the missing-data mechanism is characterized as \textit{missing completely at random (MCAR)}. In contrast, \textit{missing at random (MAR)} indicates that the distribution of $M$ depends only on $Y_{obs}$. The mechanism is called \textit{not missing at random (NMAR)} when the distribution of $M$ depends on $Y_{mis}$.

When data are MCAR or MAR, the maximum likelihood estimates for the data generating model parameters do not change when $M$ is taken into account~\cite{little2014statistical}. In these cases, missing-data mechanism is said to be \textit{ignorable}. When data are NMAR, however, it becomes more effective for parameter estimation to consider the joint likelihood of the data generating model and the missingness-encoding model. In a Bayesian framework, the missingness-encoding model may be represented as a probabilistic model; this model may be coupled with a arbitrary data generating model. When the missing-data mechanism is non-ignorable, we hypothesize that identifying the mechanism correctly will improve inference in the joint data generating model.
\begin{table*}[t!]
\caption{{\bf Six common additive exponential dispersion models.} Gaussian, gamma, inverse Gaussian, Poisson, binomial, and negative binomial distributions in additive EDM form. For the gamma distribution, $a$ and $b$ refer to the shape and rate parameters, respectively.}
\label{table:edm}
% \vskip -0.35in
\begin{center}
\begin{sc}
\renewcommand{\arraystretch}{1.75}
\begin{tabular}{llcccc}
\hline
distribution & {} & $\theta$ & $\kappa$ & $\Psi(\theta)$ & $h(x,\kappa)$ \\
\hline
Gaussian & $N(\mu,\sigma^2)$ & $\mu/\sigma^{2}$ & $\sigma^2$ & $\theta^2/2$ & $-\frac{x^2}{2\kappa} -\frac{1}{2}\log(2\pi \kappa)$ \\
Gamma & $Ga(a,b)$ & $-b$ & $a$ & $-\log(-\theta)$ & $(\kappa-1)\log x-\log\Gamma(\kappa)$ \\
Inv. Gaussian & $IG(\mu,\lambda)$ & $-\frac{\lambda}{2\mu^{2}}$ & $\sqrt{\lambda}$ & $-\sqrt{-2\theta}$ & $\frac{-\kappa^2}{2x}+\log(\kappa) - \frac{1}{2}\log(2\pi x^{3})$ \\
Poisson & $Po(\lambda)$ & $\log \lambda$ & $1$ & $e^{\theta}$ & $x\log\kappa - \log\Gamma(x+1)$ \\
Binomial & $Bi(r,p)$ & $\log(\frac{p}{1-p})$ & $r$ & $\log(1+e^\theta)$ & $\log\binom{\kappa}{x}$ \\
Neg. Binomial & $NB(r,p)$ & $\log p$ & $r$ & $-\log(1-e^\theta)$ & $\log\binom{x + \kappa -1}{x}$ \\
\hline
\end{tabular}
\end{sc}
\end{center}
\vskip -12pt
\end{table*}

In the theoretical machine learning literature, the missing-data problem is often discussed within the limited attribute observability framework~\cite{birkendorf1998restricted, cesa2011efficient, hazan2012linear, kukliansky2015attribute} where it is assumed that the learner controls which attributes it may observe. These models address a variety of data generating models. \citet{chechik2008max} considered the situation where the learner does not control attribute observability in the linear regression setting. Similarly, \citet{hazan2015classification} proposed a non-probabilistic algorithm for the classification problem under the low-rank assumption and extreme sparsity. Probabilistic models for low-rank approximation of extremely sparse matrices are abundant, and include probabilistic matrix factorization (PMF)~\cite{salakhutdinov2011probabilistic}, non-negative matrix factorization (NMF)~\cite{lee1999learning}, and their variants.

On the applied side, a motivating missing-data problem with extreme sparsity is \emph{collaborative filtering}---creating a predictive ranking of items for each user given observations of users' preferences---where capturing the missing-data pattern is crucial for accurate ranking~\cite{hu2008collaborative}. State-of-the-art probabilistic collaborative filtering models are based on either PMF or Poisson factorization (PF), the probabilistic counterpart of NMF~\cite{cemgil2009bayesian}. For instance, weighted matrix factorization (WMF)~\cite{hu2008collaborative} is a PMF model fit to binarized data with a heteroscedastic variance term. The exposure matrix factorization model (ExpoMF) uses a WMF model conditioned on a Bernouilli exposure matrix to capture the missing-data pattern~\cite{liang2015modeling}. Another successful implicit feedback model is hierarchical Poisson factorization (HPF)~\cite{gopalan2013scalable}. Building on the HPF, hierarchical compound Poisson factorization (HCPF) uses PF to encode the missing-data pattern and extends the PF structure with a general additive exponential dispersion model to generate data~\cite{basbug2016hierarchical}. HCPF is a flexible model that can be used to factorize continuous real-valued or non-negative data as well as non-negative discrete data.

The increasing popularity of PF-based methods in collaborative filtering can be attributed to two factors. First, the Gamma-Poisson distributions of PF are conjugate whereas the Gaussian-Bernoulli distributions of PMF models are not. Thus, probabilistic inference in PF-based models for large scale problems is more straightforward and computationally tractable \cite{gopalan2013scalable,basbug2016hierarchical}. Second, in HPF and HCPF, hierarchical structure models the user activity and the item popularity in a natural way. In particular, \citet{gopalan2013scalable} show that the heavy tail Gamma priors in HPF accurately capture the user behavior and item popularity by using posterior predictive checks.

An equally important but more complex aspect of implicit feedback models is the missing-data mechanism. \citet{marlin2009collaborative} give ample empirical evidence motivating the need for, and showing the performance benefits of, explicitly modeling NMAR missing-data mechanisms in collaborative filtering models.  Recommendation systems have largely driven the development of collaborative filtering models that include NMAR data; however, the problem of NMAR data exist in a broad range of analytic tasks that have been underdeveloped up to now. Motivated by these two observations, we develop an explicit NMAR missing data mechanistic model coupled with an arbitrary generative model framework. We show the benefits of including an explicit missing-data mechanistic model on three specific generative model tasks: mixture models for clustering, linear regression for prediction, and latent factor models for matrix factorization.

We start with data generating models with an additive exponential dispersion model output. This large collection of models includes Gaussian, gamma, binomial mixture models, PMF, HPF, and linear regression models among others. We explore the relationship between the data generating model and the missingness-encoding model, and we identify the need for a missing-data mechanism that can capture a heteroscedastic relationship between observations and the missing-data pattern. More specifically, we empirically show that the variance of an observation $y_{ij}$ is a function of probability of missingness of that observation, $Pr(M_{ij})$, in three large data sets. To address this issue, we propose the coupled compound Poisson factorization (CCPF) framework as a missing-data mechanism for NMAR data with extreme sparsity. We prove that the CCPF model reduces to the data generating model when the missingness-encoding model is ignorable. The first implication of this result is that it is sufficient to update the parameters of the data-generating model using the non-missing entries only. The second implication is that the statistician is able to describe the data generating model ignoring the possible impact of the missingness-encoding model on heteroscedasticity. When heteroscedasticity exists in the data, the missingness-encoding model within the CCPF framework will accurately capture and control for this complexity in the analytic task.

%% maybe another paragraph here
\begin{figure*}[t!]
\hfill
\subfigure[]{\includegraphics[width=8cm]{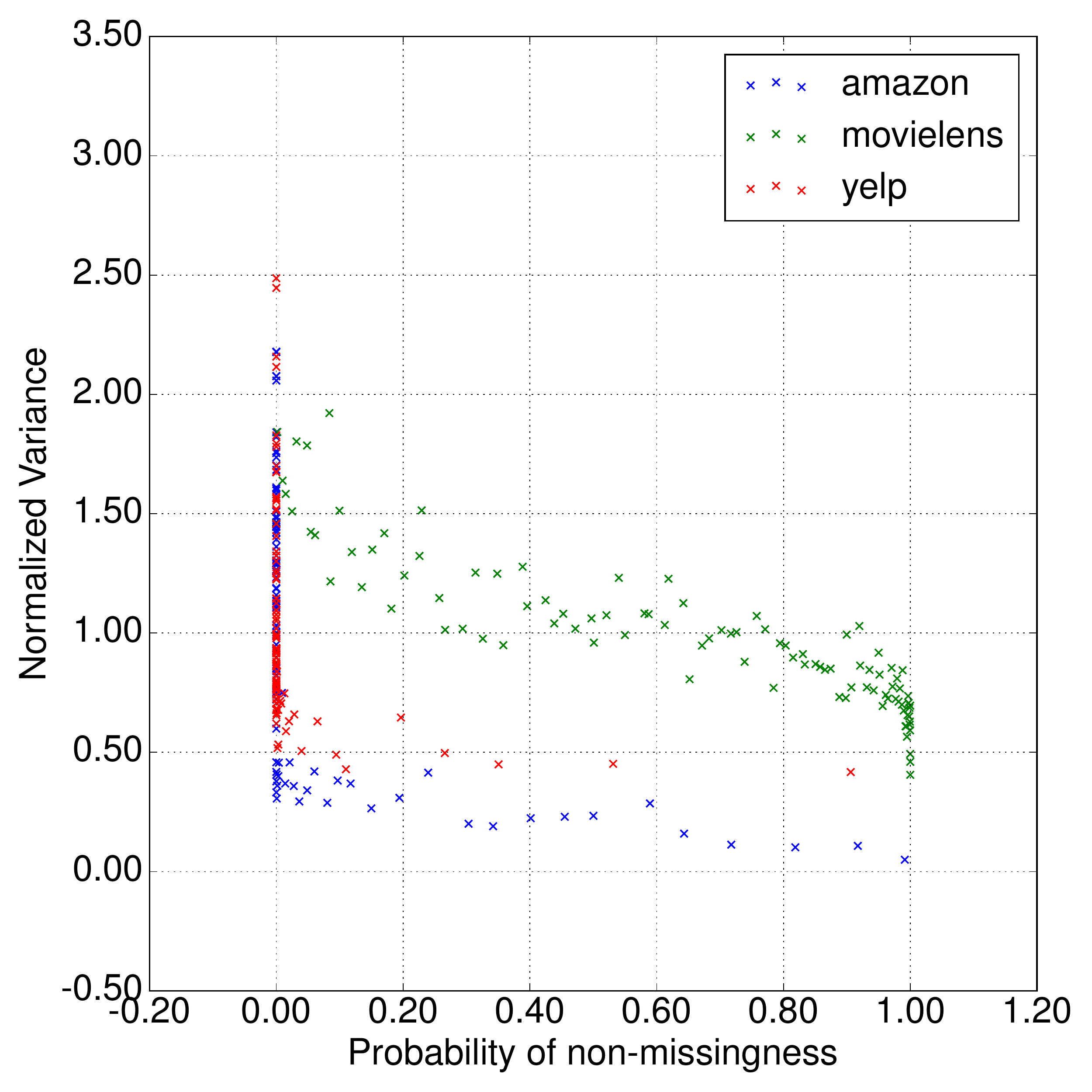}\label{fig:var_vs_quantiles}}
\hfill
\subfigure[]{\includegraphics[width=8cm]{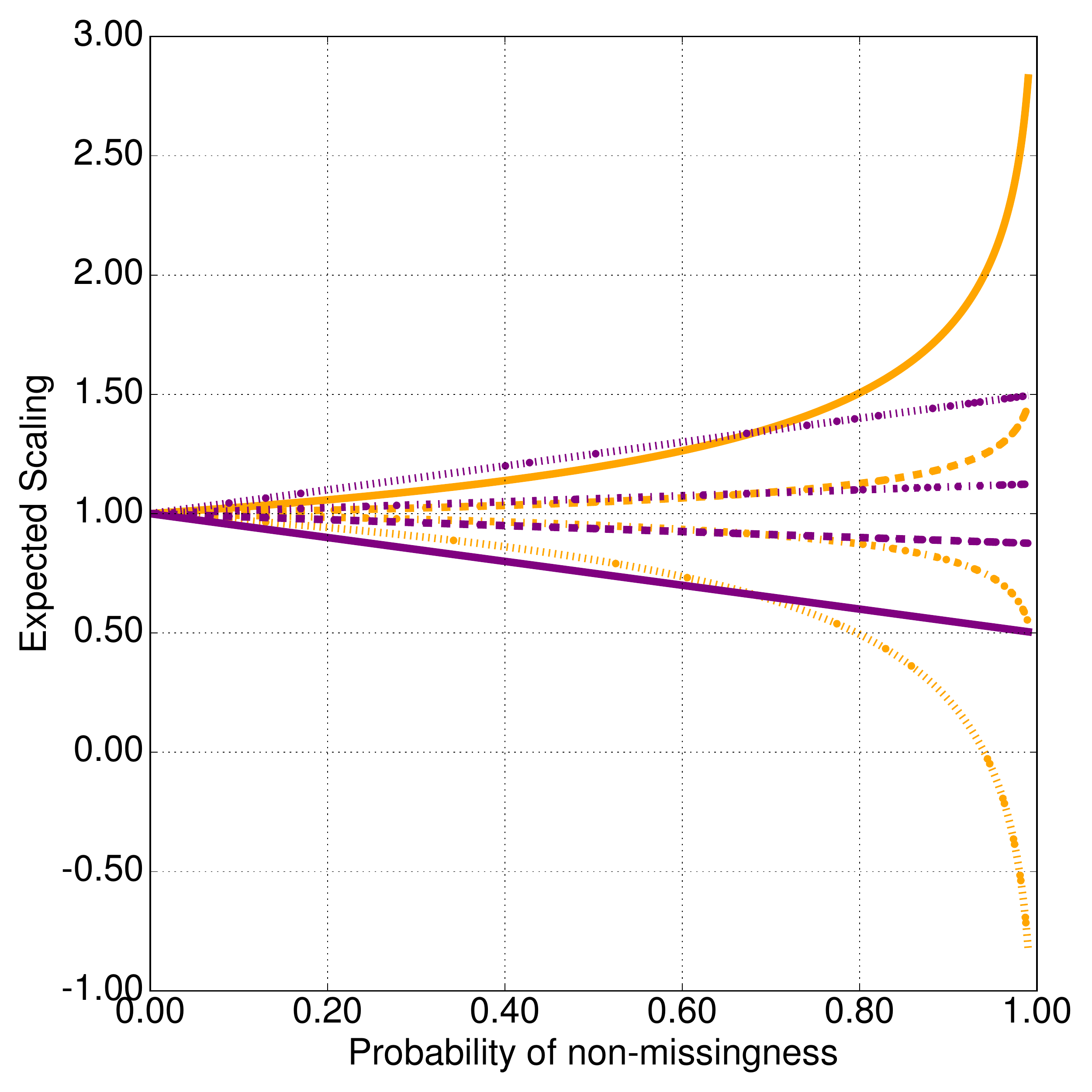}\label{fig:phi}}
\hfill
\caption{{\bf Exploratory analysis for heteroscedasticity.} a) Scatter plots of the normalized variance of the residuals of test data sets after fitting a PMF model to non-missing entries with respect to qunatiles of the probability of non-missingness calculated by fitting an HPF model to the binarized full matrix for \textit{amazon} (blue) \textit{movielens} (green) and \textit{yelp} (red) data sets, b) expected scaling versus probability of non-missingness for $c$ taking values $0.5$ (solid), $0.125$ (dashed), $-0.125$ (dash-dot), $-0.5$ (dotted) orange for exponential linkage function and purple for linear linkage function.}
\end{figure*}
% \begin{figure*}[t!]
%     \begin{center}
%       \begin{subfigure}[h]{0.45\textwidth}
%         \includegraphics[width=\textwidth]{prob_of_nm.pdf}
%         \caption{}
%         \label{fig:var_vs_quantiles}
%       \end{subfigure}
%       ~
%       \begin{subfigure}[h]{0.45\textwidth}
%         \includegraphics[width=\textwidth]{phi.pdf}
%         \caption{}
%         \label{fig:phi}
%       \end{subfigure}
%     \caption{{\bf Exploratory analysis for heteroscedasticity.} a) Scatter plots of the normalized variance of the residuals of test data sets after fitting a PMF model to non-missing entries with respect to qunatiles of the probability of non-missingness calculated by fitting an HPF model to the binarized full matrix for \textit{amazon} (blue) \textit{movielens} (green) and \textit{yelp} (red) data sets, b) expected scaling versus probability of non-missingness for $c$ taking values $0.5$ (solid), $0.125$ (dashed), $-0.125$ (dash-dot), $-0.5$ (dotted) orange for exponential linkage function and purple for linear linkage function.}
%   \end{center}
%   \vspace{-12pt}
% \end{figure*}

\section{Additive Exponential Dispersion Models}
\label{sec:edm}
We start by describing the family of \textit{additive exponential dispersion models} and its defining characteristic, additivity. Exponential dispersion models (EDM) have been primarily used as error distributions for generalized linear models \cite[for a comprehensive treatment of the theory of EDMs, see][]{jorgensen1997theory}. Additive EDMs, a subfamily of EDMs, includes Gaussian, gamma, inverse Gaussian, Poisson, binomial, and negative binomial distributions, among others (see Table~\ref{table:edm}). Following \citet{basbug2016hierarchical}, we define additive EDMs as follows.
\begin{defn} A family of distributions $\mathcal{F}_{\Psi}= \left \{ p_{(\Psi,\theta,\kappa)} \mid \theta \in \Theta  = dom(\Psi) \subseteq \mathbb{R}, \kappa \in \mathbb{R}_{++} \right \}$ is called an \emph{additive exponential dispersion model} if
\begin{align*}
p_{(\Psi,\theta,\kappa)}(x) &= \exp( x\theta - \kappa\Psi(\theta) + h(x,\kappa))
\label{density_x}
\end{align*}
where $\theta$ is the natural parameter, $\kappa$ is the dispersion parameter, $\Psi(\theta)$ is the base log-partition function, and $h(x,\kappa)$ is the base measure.
\end{defn}

With the definition above, we see that the sum of additive EDMs sharing $\Psi$ and $\theta$ has the same base log-partition function and natural parameter. The following theorem makes this statement concrete:
\begin{theorem}~\cite{jorgensen1997theory}
Let $X_{1}\dots X_{M}$ be a sequence of additive EDMs such that $X_{j} \sim p_{\Psi}(x;\theta,\kappa_{j})$, then $X_{+} = \sum_{j}X_{j} \sim p_{\Psi}(x;\theta,\sum_{j}\kappa_{j})$.
\label{th:edm}
\end{theorem}

We will exploit this property to provide theoretical justification of compounding in our coupling framework. More importantly, inference in compound Poisson additive EDMs is relatively straightforward due to the availability of the conditional and marginal densities~\cite{basbug2016hierarchical}.

\section{Coupling Framework}
\label{sec:coupling}
Let $y_{ij} \sim p_{\Psi}(\theta_{ij},\kappa)Pr(\theta_{ij} \mid \boldsymbol{\vartheta})$ be an arbitrary data generating model such that $p_{\Psi}(\theta_{ij},\kappa)$ is an additive EDM with dispersion parameter $\kappa$, and $Pr(\theta_{ij} \mid \boldsymbol{\vartheta})$ is a conjugate prior hierarchical model with parameters $\boldsymbol{\vartheta}$. Examples of such data generating models include Gaussian, Poisson, gamma, and binomial mixture models, PMF, HPF, and linear regression models.

\begin{table*}[t!]
\caption{{\bf Data set statistics} Number of samples, attributes, non-missing entries, and the ratio of missing entries to the total number of entries (sparsity), the maximum and minimum observation values.}
\label{table:datasets}
\vskip -12pt
\begin{center}
\begin{sc}
\begin{tabular}{lrrrrrr}
\hline
{data set}   & \# samples & \# attributes &  sparsity & \# non-missing &           min  & max          \\
\hline
amazon       &    256,059 &        74,258 &  0.999970 &        568,454 &              1 & 5            \\
movielens    &    247,753 &         9,732 &  0.990723 &     22,369,101 &              1 & 10           \\
netflix      &    480,189 &        17,770 &  0.988224 &    100,483,024 &              1 & 5            \\
yelp         &    552,339 &        77,079 &  0.999948 &      2,225,213 &              1 & 5            \\
wordpress    &     86,661 &        78,754 &  0.999915 &        581,508 &              1 & 1013         \\
geuvadis     &      9,358 &           462 &  0.462121 &      2,325,461 &          -19.3 & 12.6       \\
\hline
\end{tabular}
\end{sc}
\end{center}
\vskip -12pt
\end{table*}
First, we investigate the relationship between the missing-data pattern and the data-generating model. We fit a PMF model to three different data sets with MCAR assumption, i.e., only the non-missing entries are sampled during training. We also fit an HPF model to the binarized full matrix for each of the data sets. For a test set of non-missing entries, we calculate the residuals under PMF ($y_{ij}-\hat{y}_{ij}$) and the probability of non-missingness under HPF, $Pr(M_{ij}=0)$. We discretize data into $100$ equal-sized bins based on sample quantiles of probability of non-missingness, $Pr(M_{ij}=0)$. For each bin, we calculate the variance of the residuals under PMF, $Var(y_{ij}-\hat{y}_{ij})$. Fig.~\ref{fig:var_vs_quantiles} shows the scatter plot of variance within each bin and the quantiles of probability of non-missingness for \textit{amazon}, \textit{movielens} and \textit{yelp} data sets. We clearly see that there is a linear/exponential relationship between $Var(y_{ij}-\hat{y}_{ij})$ and $Pr(M_{ij}=0)$ in all three data sets. We also test for heteroscedasticity by fitting a double generalized linear model (DGLM)~\cite{dunn2012dglm} to the residuals using the probability of non-missingness as the regressor. We conclude that there is strong heteroscedasticity in all three data sets (max $p \leq 6\times 10^{-17}$).

To address the heteroscedastic structure in data analysis with missing observations, we present a framework for NMAR missing-data mechanism that can handle large-scale, extremely sparse data with two additional desirable properties: i) compatibility with the low-rank assumption for the missing-data pattern, $M$, ii) convergence to the homoscedastic model $y_{ij} \sim p_{\Psi}(\theta_{ij},\kappa)$ when $y_{ij}$ is almost surely missing ($Pr(M_{ij}=0) \rightarrow 0$).

Let the missingness-encoding model be an HPF~\cite{gopalan2013scalable,basbug2016hierarchical}. In HPF, an entry, $y_{ij}$, is missing when a draw from the model, $n_{ij}$, is zero; otherwise, $y_{ij}$ = $n_{ij}$. HCPF has a more flexible observation model, i.e., $y_{ij} \sim p_{\Psi}(\theta,n_{ij}\kappa)$, when $n_{ij} \neq 0$. However, HCPF assumes that the natural parameter $\theta_{ij}$ is the same for all observations; therefore, the data-generating model is a fixed distribution and not a complex structure such as a mixture model or a regression model. In our framework, we assume an arbitrary data generating model $y_{ij} \sim p_{\Psi}(\theta_{ij},\kappa)Pr(\theta_{ij} \mid \boldsymbol{\vartheta})$ whose output is an additive EDM.

We couple the missingness-encoding and the data generating models by scaling the dispersion parameter $\kappa$ with a linkage function $\phi(n_{ij})$. The full generative model of the coupled compound Poisson factorization (CCPF) is as follows:
\newpage
\begin{itemize}
\item For each row $i = 1,\dots,C_{I}$
\begin{enumerate}
\item Sample activity $r_{i} \sim Ga(\rho, \varrho)$
\item For each component $k$, sample factor weight $u_{ik} \sim Ga(\eta,r_{i})$
\end{enumerate}
\item For each column $j= 1,\dots,C_{J}$
\begin{enumerate}
\item Sample popularity $w_{j} \sim Ga(\upsilon, \nu)$
\item For each component $k$, sample factor weight $v_{jk} \sim Ga(\zeta,w_{j})$
\end{enumerate}
\item For each $i$ and $j$
\begin{enumerate}
\item Sample interaction variable\\
$n_{ij} \sim Po(\Lambda_{ij} = \sum_{k}u_{ik}v_{jk})$
\item if $n_{ij}$ is $0$, then $y_{ij}$ is missing
\item else
\begin{enumerate}
\item sample natural parameter $\theta_{ij}$ from the data generating model $Pr(\theta_{ij} \mid \boldsymbol{\vartheta})$
\item sample observation\\
$y_{ij} \sim p_{\Psi}(\theta_{ij},\phi(n_{ij}) \kappa)$.
\end{enumerate}
\end{enumerate}
\end{itemize}

\begin{algorithm}[t!]
   \caption{SVI for CCPF}
   \label{alg:svi}
\begin{algorithmic}
   \STATE {\bfseries Initialize:} Hyper parameters $c,\eta,\zeta,\rho,\varrho,\upsilon,\nu$
   \STATE Set the learning rates $t_{i} = t_{j} = t_{0}$ and $\tau_{i} = \tau_{j} = t_{0}^{-\xi}$
   \REPEAT
   \STATE Sample an observation $y_{ij}$ uniformly
   \STATE Calculate $ \Lambda_{ij} = \sum_{k} \alpha^{u}_{ik} \alpha^{v}_{jk} / (\beta^{u}_{ik} \beta^{v}_{jk})$
   \IF{$y_{ij}$ is missing}
   \STATE $q(n_{ij}) = \delta_{0}$
   \ELSE
   \STATE Calculate the sufficient statistic $E[\Psi(\theta_{ij})]$ from the data-generating model
   \STATE Update the local variational parameters of the missingness-encoding model
   \begin{align*}
     q(n_{ij} = n) \propto \exp&\left \{-\kappa \phi(n) E[\Psi(\theta_{ij})]\right.\\
     &\left.+ h(y_{ij},\phi(n)\kappa) \right \}\frac{\Lambda_{ij}^{n}}{n!}\\
     \varphi_{ijk} \propto \exp& \left \{ \Psi(\alpha^{u}_{ik}) - \log \beta^{u}_{ik}\right.\\
     &\left.+\Psi(\alpha^{v}_{jk}) - \log \beta^{v}_{jk} \right \}
   \end{align*}
   \STATE Update the variational parameters of the data generating model using $E[\phi(n_{ij})]$
   \ENDIF
   \STATE Update global variational parameters
    \begin{align*}
    \alpha^{r}_{i} &= (1-\tau_{i})\alpha^{r}_{i} + \tau_{i}\left (\rho + K\eta \right )\\
    \alpha^{w}_{j} &= (1-\tau_{j})\alpha^{w}_{j} + \tau_{j}\left (\upsilon + K\zeta\right )\\
    \alpha^{u}_{ik} &= (1-\tau_{i})\alpha^{u}_{ik} + \tau_{i}\left (\eta + C_{I} E[n_{ij}] \varphi_{ijk}\right )\\
    \alpha^{v}_{jk} &= (1-\tau_{j})\alpha^{v}_{jk} + \tau_{j}\left (\zeta + C_{U} E[n_{ij}] \varphi_{ijk}\right )\\
    \beta^{r}_{i} &= (1-\tau_{i})\beta^{r}_{i} + \tau_{i}\left (\varrho + \sum_{k} \frac{\alpha^{u}_{ik}}{\beta^{u}_{ik}}\right )\\
    \beta^{w}_{j} &= (1-\tau_{j})\beta^{w}_{j} + \tau_{j}\left (\nu + \sum_{k} \frac{\alpha^{v}_{jk}}{\beta^{v}_{jk}}\right )\\
    \beta^{u}_{ik} &= (1-\tau_{i})\beta^{u}_{ik} + \tau_{i}\left (\frac{\alpha^{r}_{i}}{\beta^{r}_{i}} + C_{I} \frac{\alpha^{v}_{jk}}{\beta^{v}_{jk}}\right )\\
    \beta^{v}_{jk} &= (1-\tau_{j})\beta^{v}_{jk} + \tau_{j}\left (\frac{\alpha^{w}_{j}}{\beta^{w}_{j}} + C_{U} \frac{\alpha^{u}_{ik}}{\beta^{u}_{ik}}\right )
    \end{align*}
   \STATE Update user learning rate $t_{i} = t_{i} + 1$ and  $\tau_{i} = t_{i}^{-\xi}$
   \STATE Update item learning rate $t_{j} = t_{j} + 1$ and  $\tau_{j} = t_{j}^{-\xi}$
   \STATE (Optional) Update hyper parameters $c$ and $\kappa$
   \UNTIL{validation set log likelihood converges}
\end{algorithmic}
\end{algorithm}
First, we summarize the model assumptions behind HPF when used to capture the missing-data pattern. HPF is an extension of Poisson factorization (PF) which is essentially a non-negative matrix factorization model. Imagine we have a movie ratings data set, for which an entry is missing when a user has not rated a movie. In PF setting, the missing-data pattern is approximated with an interaction matrix where each entry is a latent Poisson random variable ($n_{ij}$ for the $i^{th}$ user and $j^{th}$ movie). The interaction variable, $n_{ij}$, is the sum of $k$ interaction contributions, $n_{ijk}$, which are themselves are Poisson distributed ($n_{ijk}\sim Po(u_{ik}v_{jk})$). The low rank assumption $k << C_{I}$ implies that there exist latent groups of users and the factor weight $u_{ik}$ models the membership of $i$ to $k^{th}$ group. Different than PF, HPF has another random variable $r_{i}$ modeling how active the user $i$ is (i.e. how many movies user $i$ has rated). If a user is active, then the factor weight $u_{ik}$ and the interaction contribution $n_{ijk}$ is adjusted accordingly. Similarly, there exist latent group of movies ($k << C_{J}$) where the factor weight $v_{jk}$ models the membership of movie $j$ to $k^{th}$ genre. The popularity variable $w_{j}$ controls how many times the $j^{th}$ movie has been rated. The key point is that the interactions between groups of users and the groups of movies are non-negative. A group of users can only `not interact' with a movie genre, hence $n_{ijk}$ cannot be negative. This is different than performing matrix factorization on movie ratings where a user group may dislike a movie genre and down-vote. Furthermore, the latent factors of movies or users in terms interaction can be vastly different than the latent factors in terms preference. This brings us to the data generating model. In CCPF, we have the ability to choose an arbitrary data-generating model $Pr(\theta_{ij} \mid \boldsymbol{\vartheta})$ for the movie ratings, perhaps another matrix factorization model. We model the relationship between the missingness-encoding model and the data-generating model via the linkage function $\phi(n_{ij})$.

Second, we note that the probability distribution of the missing-data pattern, $M$, is given by
\begin{align*}
Pr(M_{ij} = 1) = Pr(y_{ij} \text{ is missing}) = e^{-\Lambda_{ij}}.
\end{align*}
In the generative model, we have the linkage function $\phi(n_{ij})$ scaling the dispersion parameter, $\kappa$. The conditional mean and variance of $y_{ij}$ are given by
\begin{align*}
E[y_{ij}\mid n_{ij}] &= \phi(n_{ij}) \kappa \Psi'(\theta_{ij})\\
Var(y_{ij}\mid n_{ij}) &= \phi(n_{ij}) \kappa  \Psi''(\theta_{ij}).
\end{align*}

The data generating model with Gaussian observations is a special case of this framework because $\theta_{ij}$ is also additive. Thus, we can use separate scaling functions $\phi_{\mu}(n_{ij})$ and $\phi_{\sigma}(n_{ij})$ that result in
\begin{align*}
E[y_{ij}\mid n_{ij}] &= \phi_{\mu}(n_{ij}) \mu_{ij}\\
Var(y_{ij}\mid n_{ij}) &= \phi_{\sigma}(n_{ij}) \sigma^2.
\end{align*}

\paragraph{Ignorable missing-data mechanism:} When $\phi(n_{ij}) = \phi^{I}(n_{ij}) = 1$, the data generating model is decoupled from the missingness-encoding model. This is equivalent to training the two models separately.

\paragraph{Linear missing-data mechanism:} To obtain a linear relationship between the probability of missingness and the dispersion parameter, we set the linkage function as $\phi(n_{ij}) = \phi^{L}(n_{ij}) = 1 - c + c (-1)^{n_{ij}+1}$. The expectation of $\phi^{L}(n_{ij})$ under the zero truncated Poisson (ZTP) distribution $n_{ij}\sim ZTP(\Lambda_{ij})$ is given by
\begin{align*}
E\left[\phi^{L}(n_{ij}) = 1 - c + c (-1)^{n_{ij}+1}\right] =& 1-c + ce^{-\Lambda_{ij}}.
\end{align*}
In Fig.~\ref{fig:phi}, the purple lines indicate the expected scaling as the probability of non-missingness changes between $0$ and $1$. As seen in the figure, when $c > 0$, there is an inverse linear relationship between the two. In other words, as the probability of non-missingness increases the expected dispersion decreases linearly. This is compatible with the empirical findings in Fig.~\ref{fig:var_vs_quantiles}.

\paragraph{Exponential missing-data mechanism:} Another possible linkage function is exponential. Setting $\phi(n_{ij}) = \phi^{E}(n_{ij}) = 1 - c + c n_{ij}$ with $c>0$ implies that, as the probability of non-missingness increases, we expect a greater dispersion. Fixing $c=1$, we get a standard compound Poisson additive EDM model~\cite{basbug2016hierarchical}. When $c<0$, there is an inverse relationship between the probability of non-missingness and the dispersion. The expectation of $\phi^{E}(n_{ij})$ under the ZTP is given by
\begin{align*}
E\left[\phi^{E}(n_{ij}) = 1 + c(n_{ij}-1)\right] =& 1-c + c\frac{e^{\Lambda_{ij}}\Lambda_{ij} }{e^{\Lambda_{ij}}-1}.
\end{align*}
In Fig.~\ref{fig:phi}, the orange lines show the expected scaling with respect to the probability of non-missingness for different values of $c$. When $c>0$, the expected scaling increases exponentially as the probability of non-missingness increases. Expected scaling decreases exponentially, when $c < 0$. Since dispersion is always positive, one needs to be careful in choosing $c$. Exponential decay with a small $c$ might be a better choice than the linear relationship for certain data sets as seen in Fig.~\ref{fig:var_vs_quantiles}.

Another takeaway from Fig.~\ref{fig:phi} is that the expected scaling converges to $1.0$ as the probability of non-missingness becomes $0$. With the following theorem, we provide a stronger result.

\begin{theorem}
\label{th:convergence}
Let $N$ be a zero-truncated Poisson random variable with parameter $\Lambda$, and let $Y_{1}\dots Y_{N}$ be i.i.d. additive EDM random variables such that $Y_{j} \sim p_{\Psi}(\theta, c\kappa)$, where $0<c<1$. Let $Y_{0} \sim p_{\Psi}(\theta, (1-c)\kappa)$ be another random variable with the same log-partition function $\Psi$ and the natural parameter $\theta$. Define the linear linkage function as $\phi^{L}(N) \doteq 1 - c + c (-1)^{N+1}$ and the exponential linkage function as $\phi^{E}(N) \doteq 1 - c + c N$. Then the probability densities of $Y^{L}_{+}~\doteq~Y_{0}~+~\sum_{n=1}^{N}~(-1)^{n+1}Y_{n}$ and $Y^{E}_{+}~\doteq~\sum_{n=0}^{N} Y_{n}$ are given by
\begin{align*}
Y^{L}_{+} &\sim p_{\Psi}(\theta, \phi^{L}(N)\kappa) \\
Y^{E}_{+} &\sim p_{\Psi}(\theta, \phi^{E}(N)\kappa).
\end{align*}
Furthermore, both $Y^{L}_{+}$ and $Y^{E}_{+}$ converge to $Y \sim p_{\Psi}(\theta, \kappa)$ in distribution as $\Lambda$ goes to zero; that is
\begin{align*}
Y^{L}_{+} &\overset{D}{\rightarrow} Y \;\;\text{as} \;\;\Lambda\rightarrow 0\\
Y^{E}_{+} &\overset{D}{\rightarrow} Y \;\;\text{as} \;\;\Lambda\rightarrow 0.
\end{align*}
\end{theorem}
Theorem~\ref{th:convergence} implies that the distribution of an observation $y_{ij}$ that is missing almost surely ($\Lambda\rightarrow 0$) converges to the homoscedastic observation model, i.e., $p_{\Psi}(\theta_{ij}, \kappa)$. From the model building aspect, this is a useful property, since we can design the data-generating model independently and simply plug in the missingness-encoding model to capture any suspected heteroscedasticity. Another corollary of this theorem is that the parameters of data generating model $Pr(\theta_{ij} \mid \boldsymbol{\vartheta})$ can be estimated using the non-missing entries only, as would be done for homoscedastic model.

For computationally tractable learning, we use stochastic variational inference (SVI)~\cite{hoffman2013stochastic}, which minimizes the lower bound on the expected posterior log likelihood under a variational distribution. The mean field variational distribution for CCPF is given by
\begin{align*}
&q(r_{i}\mid \alpha^{r}_{i}, \beta^{r}_{i})
q(u_{ik}\mid \alpha^{u}_{ik},\beta^{u}_{ik})
q(w_{j}\mid \alpha^{w}_{j},\beta^{w}_{j})\\
&q(v_{jk}\mid \alpha^{v}_{jk},\beta^{v}_{jk})
q(\hat{n}_{ij}\mid \boldsymbol{\varphi_{ij}})
q(n_{ij})q(\theta_{ij})q(\boldsymbol{\vartheta}),
\end{align*}
where the distributions of the variational approximation are as in the generative distribution. For gamma distributions, $\alpha$ is the shape and $\beta$ is the rate parameter. We make use of the multinomial representation of Poisson factors as in~\cite{cemgil2009bayesian,gopalan2013scalable,basbug2016hierarchical}. To update the variational parameters of the data generating model, we need the sufficient statistic $E[\phi(n_{ij})]$. Similarly, we need the sufficient statistics $E[\Psi(\theta_{ij})]$ to update the variational parameters of the missingness-encoding model. The SVI algorithm for the general CCPF framework is summarized in Algorithm~\ref{alg:svi}.

\section{Experiments}
We apply the CCPF framework to three classes of data generating models: mixture models, matrix factorization, and linear regression. For each model, we compare to CCPF to data-generating models that ignore the missing-data mechanism. In each case, the CCPF approach outperforms the comparisons in terms of the log likelihood of held-out non-missing entries.

\begin{table*}[t!]
\caption{{\bf Non-missing test set log likelihood for mixture models.} Log likelihood per non-missing test entry for clustering performed across samples and attributes. GMM and PMM stand for Gaussian mixture model and Poisson mixture model, respectively. CCPF-GMM and CCPF-PMM are the variants of these two models in our framework.}
\label{table:tll-mixture}
\vskip -12pt
\begin{center}
\begin{sc}
\begin{tabular}{llllllll}
{}       &       axis & sampling & movielens &     amazon &   netflix &     yelp   &  geuvadis \\
\hline
GMM      & sample    & Nonzero &    -2.082 &     -1.549 &    -1.427 &    -1.709  &    -2.139 \\
GMM      & attribute & Nonzero & \bf-2.064 &     -1.667 &    -1.491 &    -1.637  &    -2.173 \\
\bf CCPF-GMM & sample    &    Full &    -2.121 &  \bf-1.487 & \bf-1.419 &    -1.708  & \bf-2.012 \\
\bf CCPF-GMM & attribute &    Full &    -2.102 &     -1.669 &    -1.480 & \bf-1.615  &    -2.083 \\
PMM      & sample    & Nonzero &    -3.554 &     -1.793 &    -3.682 &    -1.912  &       N/A \\
PMM      & attribute & Nonzero &    -3.674 &     -1.886 &    -1.918 &    -1.793  &       N/A \\
\bf CCPF-PMM & sample    &    Full &    -3.713 &     -1.789 &    -3.315 &    -1.889  &       N/A \\
\bf CCPF-PMM & attribute &    Full &    -3.655 &     -1.913 &    -2.001 &    -1.801  &       N/A \\
\hline
\end{tabular}
\end{sc}
\end{center}
\vskip -12pt
\end{table*}

Mixture models, linear regression, and matrix factorization are three useful data generating models for the analysis of high-dimensional data.  Mixture models represent the partitioning of observations into subgroups, and are used for exploratory data analysis.  Matrix factorization decomposes a matrix of observations into two lower dimensional matrices, allowing each observation from the original matrix to be represented as the weighted linear combination of a lower dimensional space.
Linear regression models the relationship between regressors and observations; in our case, the observations are multivariate with a large number of dimensions.\footnote{Note that CCPF is only applicable to a linear regression settings where we have full access to regressors, but multivariate observations are missing.  It is not applicable to data with missing covariates.}  All three of these models may be applied to the same data, depending on the objectives of the analysis.  The goal of CCPF is to account for missing data under various modeling paradigms.

\subsection{Data sets}
We analyze four user behavior data sets and one gene expression data set; Table~\ref{table:datasets} outlines the characteristics of each.
The user behavior data includes multiple ratings data sets: \textit{amazon} contains ratings of fine food~\cite{mcauley2013amateurs}, \textit{netflix} consists of movie ratings~\cite{bell2007lessons}, and \textit{yelp} comprises venue ratings; ratings for all three of these data sets range from $1$ to $5$. We also consider \textit{movielens}, another movie data set with ratings ranging from $1$ to $10$. In this data, each movie has $1129$ corresponding exogenous variables representing the association with a predetermined tag. The tag weights are calculated from the crowd sourced tag-movie association data~\cite{harper2015movielens}.  Behavior data is not always bounded---\textit{wordpress} is a social media interaction data set of users and blogs where the response is the number of likes a user had for a given blog. The \textit{geuvadis} data shows that CCPF is applicable beyond discrete user behavior data; it is a gene expression data set of 9,358 genes for each of 462 individuals.

\subsection{Experimental Details}
We held out $20\%$ and $1\%$ of the non-missing entries for testing ($\mathcal{Y}_{obs}^{test}$) and validation, respectively. Test log likelihood of the non-missing entries ($\mathcal{L}_{NM}$) under CCPF is calculated as
\begin{align*}
\mathcal{L}_{NM} &=\sum_{\mathcal{Y}_{obs}^{test}} \log \sum_{n=1}^{N_{tr}} p_{\Psi}(y_{ij}^{test};\theta_{ij},\phi(n)\kappa) ZTP(n \mid \Lambda_{ij}).
\end{align*}
To set the hyper parameters of the missingness-encoding model we followed the method presented for HCPF~\cite{basbug2016hierarchical}. We fix $K=160$, $\xi = 0.7$ and $t_{0} = 10,000$ after an empirical study on small data sets. We estimate the sparsity level from the number of non-missing entries and from the sparsity level we calculated $E[n_{ij}]$. For heavy tail Gamma priors, we set $\nu=\varrho=0.1$ and $\upsilon=\rho=0.01$.  We then set $\eta = \rho \sqrt{E[n_{ij}]/K}/\varrho$ and $\zeta = \upsilon \sqrt{E[n_{ij}]/K}/\nu$ as done in~\cite{basbug2016hierarchical}. We set $N_{tr}$ using the
expected range of $\lambda_{ij}$ and $y_{ij}$ as well as the maximum likelihood estimates of $E[\Psi(\theta_{ij})]$ and $\kappa$ under the assumption that $\theta_{ij}$ is fixed. We have an inverse Gamma prior on $\kappa$ with shape parameter $1.01$ and scale parameter $1.0$. We note that the inverse Gamma is the conjugate prior to $\kappa$ for the exponential dispersion models under the saddle-point approximation~\cite{jorgensen1997theory}. Additionally, we have a Gaussian prior with zero mean and $0.1$ standard deviation on the linkage parameter $c$. We then calculate the MAP estimates of $\kappa$ and $c$ using stochastic gradient descent with smoothed gradients by taking the average of $1000$ gradients.

\subsection{Mixture Model Results}
We first consider Gaussian mixture models (GMM) with spherical Gaussian priors and Poisson mixture models (PMM) with Gamma priors as our data-generating models. We derive SVI algorithms modified to accommodate coupling (see Appendix).

\begin{table*}[t!]
\caption{{\bf Non-missing test set log likelihood for factorization models.} PMF stands for the probabilistic matrix factorization, HPF stands for hierarchical Poisson factorization. CCPF-PMF and CCPF-HPF are the variants of these two models in our framework.}
\vskip -12pt
\begin{center}
\begin{sc}
\begin{tabular}{lrrrrrr}
{}       & sampling & movielens &     amazon &   netflix &    yelp   & wordpress \\
\hline
PMF      &  Nonzero &    -2.089 &     -1.703 &    -1.421 &    -1.952 &    -3.379 \\
HPF      &  Nonzero &    -2.104 &     -1.868 &    -1.634 &    -1.881 &    -2.146 \\
\bf CCPF-PMF &     Full & \bf-2.009 &  \bf-1.630 & \bf-1.383 & \bf-1.734 &    -3.046 \\
\bf CCPF-HPF &     Full &    -2.013 &     -1.835 &    -1.693 &    -1.878 & \bf-1.922 \\
\hline
\end{tabular}
\label{table:tll-factorization}
\end{sc}
\end{center}
\vskip -12pt
\end{table*}
We fit GMM, PMM, CCPF-GMM and CCPF-PMM to \textit{movielens}, \textit{amazon}, \textit{netflix} and \textit{yelp} data sets. We perform clustering across users (samples) and items (attributes) separately. In the continuous gene expression data set (\textit{geuvadis}), we fit GMM and CCPF-GMM to identify cluster of individuals (samples) and genes (attributes). We first take the log of the gene expression levels. We compare models performance in terms of test set log likelihood.

In Table~\ref{table:tll-mixture}, we observe that CCPF-GMM outperforms GMM in \textit{amazon}, \textit{netflix}, \textit{yelp} and \textit{geuvadis} for clustering both samples and attributes. Only in \textit{movielens}, we see GMM beating CCPF-GMM. In PMM comparisons, there is no consistent pattern. This can be explained by our previous observation that the impact of missingness-encoding model on the data-generating model is significant for the dispersion parameter but not significant for the mean term. In the case of GMM, the scaling of dispersion parameter only affects the variance term. In PMM, $\phi(n_{ij})$ scales $\lambda_{ij}$ which is the mean and the variance of the observation; therefore, coupling is more effective in GMM.

\subsection{Matrix Factorization Results}
Next, we considered two major matrix factorization models. PMF is the probabilistic counterpart of regularized SVD where the penalty terms relate to the spherical Gaussian priors on factors~\cite{salakhutdinov2011probabilistic}. HPF is another probabilistic matrix factorization model where the factor contributions are non-negative~\cite{gopalan2013scalable}. Previously, we utilized HPF for the missingness-encoding model; however, HPF can also be trained only on the non-missing entries. We derive SVI algorithm for PMF and HPF accommodating the coupling from the missingness-encoding model (see Appendix).

We fit PMF, HPF, CCPF-PMF and CCPF-HPF to \textit{movielens}, \textit{amazon}, \textit{netflix}, \textit{yelp} and \textit{wordpress} data sets. Table~\ref{table:tll-factorization} summarizes the comparison of models in terms of test set log likelihood. In ratings data sets, CCPF-PMF is the best performing algorithm with a clear margin in most cases. Similar to the mixture model analysis, we observe a more substantial improvement in PMF to CCPF-PMF transition than HPF to CCPF-HPF transition. Social media activity data set, \textit{wordpress}, exhibits a different characteristic---Poisson models outperform PMF and CCPF-PMF. This can be attributed to the highly dispersed Poisson-like response distribution.
\begin{table}[t!]
\caption{{\bf Comparison for regression models.} Log likelihood, RMSE and $R^2$ values for hierarchical linear regression models on \textit{movielens} data set. CCPF-Regr. is the variant of the regression model in our framework.}
\label{table:tll-regression}
\vskip -12pt
\begin{center}
\begin{small}
\begin{sc}
\begin{tabular}{lllll}
{}                  & sampling & TLL            &     RMSE &    $R^2$ \\
\hline
Regression          &  Nonzero &    -1.921      &    1.731 &    0.333 \\
\bf CCPF-Regr. &     Full & \bf-1.907      & \bf1.694 & \bf0.360 \\
\hline
\end{tabular}
\end{sc}
\end{small}
\end{center}
\vskip -12pt
\end{table}
\subsection{Linear Regression Results}
Finally, we considered a hierarchical linear regression as the data-generating model. Let $\boldsymbol{x_{j}}$ be the attribute vector for item $j$. We describe the data-generating model by $y_{ij} \sim N(\boldsymbol{b_{i}}^T\boldsymbol{x_{j}},\sigma^2)$ where each user, $i$, has a unique coefficient vector $\boldsymbol{b_{i}}$. We have a Gaussian prior on the coefficient vectors that is set to the maximum likelihood estimate of the coefficient vector when the response is fixed to the mean ratings across users. SVI for the resulting coupled model is given in the Appendix.

We fit the regression model as well as its coupled version to \textit{movielens} data set where the attribute vector is the tag weights discussed earlier. The goal is to capture user preferences over arbitrary tags and use this information within the collaborative filtering setting. In Table~\ref{table:tll-regression}, we observe that coupling improves the regression performance. We also note that the regression approach achieves the highest test log likelihood among mixture model and matrix factorization approaches.

\section{Conclusion}
In this work, we present the coupled compound Poisson factorization (CCPF) that models the missing-data mechanism in extremely sparse data sets by coupling a missingness-encoding model with an arbitrary data-generating model. We derive stochastic variational inference algorithm for our CCPF models and, as examples of our framework, implement instances of a mixture model, linear regression, and matrix factorization. We compare our model with the data generating models that ignores the missing-data mechanism on large scale studies and show that explicitly modeling the missing-data mechanism substantially improves test log likelihood and other metrics relevant to the analysis of interest.
\newpage

\bibliography{ccpf}
\bibliographystyle{icml2016}
\newpage
\appendix

\twocolumn[{

\section{Mixture Model}

\subsection{Gaussian Mixture Model}
Generative model can be described as
\begin{itemize}
\item For each $u = 1,\dots,C_{U}$ and for each component $k=1,\dots,K$
\begin{enumerate}
\item Sample $s_{uk} \sim N(\eta,\rho)$
\end{enumerate}
\item For each $u$ and $i$
\begin{enumerate}
\item draw $n_{ui}$ from the missingness-encoding model
\item sample $y_{ui} \sim N(\phi_{1}(n_{ui})\sum_{k}s_{uk}, \phi_{2}(n_{ui})\sigma^2)$.
\end{enumerate}
\end{itemize}
Let the variational distribution be $Q(s_{uk}) = N(a_{uk}^{s},b_{uk}^{s})$, then stochastic variational updates are
\begin{align}
a_{uk}^{s} &\leftarrow (1-t^{-\xi})a_{uk}^{s} + t^{-\xi} \left (\frac{\eta \sigma^2 +E[\phi_{1}/\phi_{2}] \rho C_{I}y_{ui} - E[\phi_{1}^2 / \phi_{2}]\rho C_{I} \sum_{j\neq k} a_{uj}^{s}}{\sigma^2 +E[\phi_{1}/\phi_{2}]\rho C_{I} } \right )\\
b_{uk}^{s} &\leftarrow (1-t^{-\xi})b_{uk}^{s} + t^{-\xi} \left ( \frac{\rho \sigma^2}{\sigma^2 +\rho C_{I} E[\phi_{1}/\phi_{2}]} \right ).
\end{align}

\subsection{Poisson Mixture Model}
Generative model can be described as
\begin{itemize}
\item For each $u = 1,\dots,C_{U}$ and for each component $k=1,\dots,K$
\begin{enumerate}
\item Sample $s_{uk} \sim Ga(\eta,\rho)$
\end{enumerate}
\item For each $u$ and $i$
\begin{enumerate}
\item draw $n_{ui}$ from the missingness-encoding model
\item sample $y_{ui} \sim Po(\phi(n_{ui})\sum_{k}s_{uk})$.
\end{enumerate}
\end{itemize}

Let the variational distribution be $Q(s_{uk}) = Ga(a_{uk}^{s},b_{uk}^{s})$, then stochastic variational updates are
\begin{align}
\varphi_{uk} &\propto \exp \left \{ \Psi(a^{s}_{uk}) - \log b^{s}_{uk} \right \}\\
a_{uk}^{s} &\leftarrow (1-t^{-\xi})a_{uk}^{s} + t^{-\xi}\left (\eta + C_{I} \varphi_{uk} y_{ui} \right )\\
b_{uk}^{s} &\leftarrow (1-t^{-\xi})b_{uk}^{s} + t^{-\xi}\left (\rho + C_{I} E[\phi]\right ).
\end{align}
}]
\newpage
\twocolumn[{
\section{Matrix Factorization}

\subsection{Probabilistic Matrix Factorization}
Generative model can be described as

\begin{itemize}
\item For each $u = 1,\dots,C_{U}$ and for each component $k=1,\dots,K$
\begin{enumerate}
\item Sample $s_{uk} \sim N(\eta,\rho)$
\end{enumerate}
\item For each $i= 1,\dots,C_{I}$ and for each component $k=1,\dots,K$
\begin{enumerate}
\item Sample $v_{ik} \sim N(\zeta,\omega)$
\end{enumerate}
\item For each $u$ and $i$
\begin{enumerate}
\item draw $n_{ui}$ from the missingness-encoding model
\item sample $y_{ui} \sim N(\phi_{1}(n_{ui})\sum_{k}s_{uk}v_{ik}, \phi_{2}(n_{ui})\sigma^2)$.
\end{enumerate}
\end{itemize}

Let the variational distribution be $Q(s_{uk}) = N(a_{uk}^{s},b_{uk}^{s})$ and $Q(v_{ik}) = N(a_{ik}^{v},b_{ik}^{v})$, then stochastic variational updates are
\begin{align}
a_{uk}^{s} &\leftarrow (1-t^{-\xi})a_{uk}^{s} + t^{-\xi} \left (\frac{\eta \sigma^2 +\rho C_{I}a_{ik}^{v} (E[\phi_{1}/\phi_{2}] y_{ui} - E[\phi_{1}^2/\phi_{2}] \sum_{j\neq k} a_{uj}^{s} a_{ij}^{v})}{\sigma^2 +E[\phi_{1}/\phi_{2}] \rho C_{I}  (b_{ik}^{v} + (a_{ik}^{v})^{2})} \right )\\
b_{uk}^{s} &\leftarrow (1-t^{-\xi})b_{uk}^{s} + t^{-\xi} \left (\frac{\rho \sigma^2}{\sigma^2 +\rho C_{I} E[\phi_{1}/\phi_{2}] (b_{ik}^{v} + (a_{ik}^{v})^{2})} \right )\\
a_{ik}^{v} &\leftarrow (1-t^{-\xi})a_{ik}^{v} + t^{-\xi} \left (\frac{\zeta \sigma^2 +\omega C_{U}a_{uk}^{s} (E[\phi_{1}/\phi_{2}]y_{ui} - E[\phi_{1}^2/\phi_{2}] \sum_{j\neq k} a_{uj}^{s} a_{ij}^{v})
}{\sigma^2 +\omega C_{U} E[\phi_{1}/\phi_{2}] (b_{uk}^{s} + (a_{uk}^{s})^{2})} \right )\\
b_{ik}^{v} &\leftarrow (1-t^{-\xi})b_{ik}^{v} + t^{-\xi} \left (\frac{\omega \sigma^2}{\sigma^2 +\omega C_{U} E[\phi_{1}/\phi_{2}] (b_{uk}^{s} + (a_{uk}^{s})^{2})} \right ).
\end{align}

\subsection{Poisson Factorization}
Generative model can be described as
\begin{itemize}
\item For each $u = 1,\dots,C_{U}$ and for each component $k=1,\dots,K$
\begin{enumerate}
\item Sample $s_{uk} \sim Ga(\eta,\rho)$
\end{enumerate}
\item For each $i= 1,\dots,C_{I}$ and for each component $k=1,\dots,K$
\begin{enumerate}
\item Sample $v_{ik} \sim Ga(\zeta,\omega)$
\end{enumerate}
\item For each $u$ and $i$
\begin{enumerate}
\item draw $n_{ui}$ from the missingness-encoding model
\item sample $y_{ui} \sim Po(\phi(n_{ui})\sum_{k}s_{uk}v_{ik})$.
\end{enumerate}
\end{itemize}

Let the variational distribution be $Q(s_{uk}) = Ga(a_{uk}^{s},b_{uk}^{s})$ and $Q(v_{ik}) = Ga(a_{ik}^{v},b_{ik}^{v})$, then stochastic variational updates are
\begin{align}
\varphi_{uik} &\propto \exp \left \{ \Psi(a^{s}_{uk}) - \log b^{s}_{uk} +\Psi(a^{v}_{ik}) - \log b^{v}_{ik} \right \}\\
a^{s}_{uk} &= (1-t_{u}^{-\xi})a^{s}_{uk} + t_{u}^{-\xi}\left (\eta + C_{I} y_{ui} \varphi_{uik}\right )\\
b^{s}_{uk} &= (1-t_{u}^{-\xi})b^{s}_{uk} + t_{u}^{-\xi}\left (\rho + C_{I} E[n_{ui}] \frac{a^{v}_{ik}}{b^{v}_{ik}}\right )\\
a^{v}_{ik} &= (1-t_{i}^{-\xi})a^{v}_{ik} + t_{i}^{-\xi}\left (\zeta + C_{U} y_{ui} \varphi_{uik}\right )\\
b^{v}_{ik} &= (1-t_{i}^{-\xi})b^{v}_{ik} + t_{i}^{-\xi}\left (\omega + C_{U} E[n_{ui}] \frac{a^{s}_{uk}}{b^{s}_{uk}}\right )
\end{align}
}]
\newpage
\twocolumn[{

\section{Linear Regression}

\subsection{Hierarchical Linear Regression}
Generative model with a exogenous covariate matrix $x_{ik}$ can be described as
\begin{itemize}
\item For each $u = 1,\dots,C_{U}$ and for each component $k=1,\dots,K$
\begin{enumerate}
\item Sample $s_{uk} \sim N(\eta,\rho)$
\end{enumerate}
\item For each $u$ and $i$
\begin{enumerate}
\item draw $n_{ui}$ from the missingness-encoding model
\item sample $y_{ui} \sim N(\phi_{1}(n_{ui})\sum_{k}s_{uk}x_{ik}, \phi_{2}(n_{ui})\sigma^2)$.
\end{enumerate}
\end{itemize}

Let the variational distribution be $Q(s_{uk}) = N(a_{uk}^{s},b_{uk}^{s})$ , then stochastic variational updates are
\begin{align}
a_{uk}^{s} &\leftarrow (1-t^{-\xi})a_{uk}^{s} + t^{-\xi} \left (\frac{\eta \sigma^2 +\rho C_{I} (E[\phi_{1}/\phi_{2}]y_{ui} - E[\phi_{1}^2/\phi_{2}]\sum_{j\neq k} a_{uj}^{s}x_{ij})x_{ik}}{\sigma^2 +\rho C_{I} E[\phi_{1}/\phi_{2}]x_{ik}^{2}} \right )\\
b_{uk}^{s} &\leftarrow (1-t^{-\xi})b_{uk}^{s} + t^{-\xi} \left (\frac{\rho \sigma^2}{\sigma^2 +\rho C_{I} E[\phi_{1}/\phi_{2}]x_{ik}^{2}} \right )
\end{align}

\section{Proof for Theorem 2}
We follow the proof of Theorem 3 in~\citep{basbug2016hierarchical}.
\begin{proof}
Let $M_{\bar{Y}}(t)$ be the MGF of $\bar{Y}\sim p_{\Psi}(x;\theta,c\kappa)$ and $\left \{(Y_{++}^{E})_{m}\right \}_{m=1}^{\infty}$, $\left \{(Y_{++}^{L})_{m}\right \}_{m=1}^{\infty}$ be sequences of random variables where $(Y_{++}^{E})_{m} = \sum_{n=1}^{N} Y_{n}$, $(Y_{++}^{L})_{m} = \sum_{n=1}^{N} (-1)^{n+1} Y_{n}$ with $N\sim ZTP(1/m)$. The MGF of $(Y_{++}^{E})_{m}$ and $(Y_{++}^{L})_{m}$ are given by
\begin{align*}
M_{(Y_{++}^{E})_{m}}(t) &= \frac{e^{M_{Y}(t)/m}-1}{e^{1/m}-1}\\
M_{(Y_{++}^{L})_{m}}(t) &= \frac{1-e^{-1/m}}{2 M_{Y}(t)} + \frac{M_{Y}(t) \sinh(1/m)}{e^{1/m}-1}
\end{align*}
Since $\lim_{m \rightarrow \infty}M_{(Y_{++}^{E})_{m}}(t) = \lim_{m \rightarrow \infty}M_{(Y_{++}^{L})_{m}}(t) = M_{\bar{Y}}(t)$, both $Y_{++}^{E} \doteq \sum_{n=1}^{N} Y_{n}$  and $Y_{++}^{L} \doteq \sum_{n=1}^{N} (-1)^{n+1} Y_{n}$ with $N\sim ZTP(\Lambda)$ converges to $\bar{Y}$ in distribution as $\Lambda$ goes to zero. Since $\bar{Y}$ and $Y_{0}$ have the same log partition function and the same natural parameter, both $Y^{E}_{+} \doteq Y_{0} + Y^{E}_{++}$  and $Y^{L}_{+} \doteq Y_{0} + Y^{L}_{++}$ converge to $Y$ in distribution.
\end{proof}
}]
\newpage

\twocolumn[{
\section{Element Distributions}
\subsection{Gaussian}
Generative model with Normal-Gamma prior can be described as
\begin{itemize}
\item draw $\mu \sim N(\eta,1 / \lambda \rho)$ and $\rho \sim Ga(\zeta,\omega)$
\item For each $u$ and $i$
\begin{enumerate}
\item draw $n_{ui}$ from the missingness-encoding model
\item sample $y_{ui} \sim N(\phi_{1}(n_{ui})\mu, \phi_{2}(n_{ui})/\rho)$.
\end{enumerate}
\end{itemize}

Let the variational distribution be $Q(\mu) = N(\hat{\mu},\hat{\sigma}^2)$  and $Q(\rho) = Ga(\hat{a},\hat{b})$, then stochastic variational updates are
\begin{align}
\hat{\mu} &\leftarrow (1-t^{-\xi})\hat{\mu} + t^{-\xi} \left ( \frac{\eta\lambda+C_{I}C_{U} y_{ui} E[\phi_{1}/\phi_{2}]}{\lambda + C_{I}C_{U} E[\phi_{1}^2/\phi_{2}]} \right )\\
\hat{\sigma}^2 &\leftarrow (1-t^{-\xi})\hat{\sigma}^2 + t^{-\xi} \left ( \frac{\hat{b}}{\hat{a}(\lambda + C_{I}C_{U} E[\phi_{1}^2/\phi_{2}])}\right )\\
\hat{a} &\leftarrow (1-t^{-\xi})\hat{a} + t^{-\xi} \left ( \zeta + \frac{1 + C_{I}C_{U}}{2} \right )\\
\hat{b} &\leftarrow (1-t^{-\xi})\hat{b} + t^{-\xi} \left (
\omega + \frac{1}{2}C_{I}C_{U} y_{ui}^2 E[1/\phi_{2}] + \frac{1}{2}\eta^2\lambda
-\hat{\mu}(C_{I}C_{U} y_{ui} E[\phi_{1}/\phi_{2}] + \eta \lambda) \right. \\
&\left . \qquad\qquad\qquad\qquad\quad +\frac{1}{2}(\hat{\mu}^2 + \hat{\sigma}^2)(C_{I}C_{U}E[\phi_{1}^2/\phi_{2}] + \lambda ) \right )
\end{align}
}]
\end{document}